\documentclass[final,5p,times,twocolumn]{elsarticle}

\usepackage{amsmath,amssymb,amsfonts}
\usepackage{hyperref}
\usepackage{cleveref}
\usepackage{algorithmic}
\usepackage{animate}
\usepackage{graphicx}
\usepackage{multirow}
\usepackage{booktabs}
\usepackage{xcolor,colortbl}
\usepackage[nolist]{acronym}
\usepackage{subcaption}

\newcommand{\ie}{i.\,e.}
\newcommand{\eg}{e.\,g.}
\newcommand{\etal}{et\,al. }
\newcommand{\etalns}{et\,al.}

\usepackage{import}

\journal{}

\begin{document}

\begin{frontmatter}



\title{A Machine Learning Framework for Automatic Prediction of Human Semen Motility}
% Automatic Video-Based Prediction of Human Semen Motility Utilising Machine Learning and Quantised Feature Representations

\author[inst1]{Sandra Ottl}

\affiliation[inst1]{organization={Chair of Embedded Intelligence for Health Care and Wellbeing, University of Augsburg, Germany}%Department and Organization
            % addressline={Eichleitnerstraße 30}, 
            % city={Augsburg},
            % postcode={86159}, 
            % state={Bavaria},
            % country={Germany}
            }
            
\affiliation[inst2]{organization={GLAM -- Group on Language, Audio, \& Music, Imperial College London, United Kingdom}}

\author[inst1]{Shahin Amiriparian}
\author[inst1]{Maurice Gerczuk}
\author[inst1,inst2]{Björn W.\ Schuller}

\begin{abstract}
In the field of reproductive health, a vital aspect for the detection of male fertility issues is the analysis of human semen quality. Two factors of importance are the morphology and motility of the sperm cells. While the former describes defects in different parts of a spermatozoon, the latter measures the efficient movement of cells. For many non-human species, so-called Computer-Aided Sperm Analysis systems work well for assessing these characteristics from microscopic video recordings but struggle with human sperm samples which generally show higher degrees of debris and dead spermatozoa, as well as lower overall sperm motility. Here, machine learning methods that harness large amounts of training data to extract salient features could support physicians with the detection of fertility issues or in vitro fertilisation procedures. In this work, the overall motility of given sperm samples is predicted with the help of a machine learning framework integrating unsupervised methods for feature extraction with downstream regression models. The models evaluated herein improve on the state-of-the-art for video-based sperm-motility prediction.
\end{abstract}

\begin{graphicalabstract}
\centering
\includegraphics[width=\textwidth]{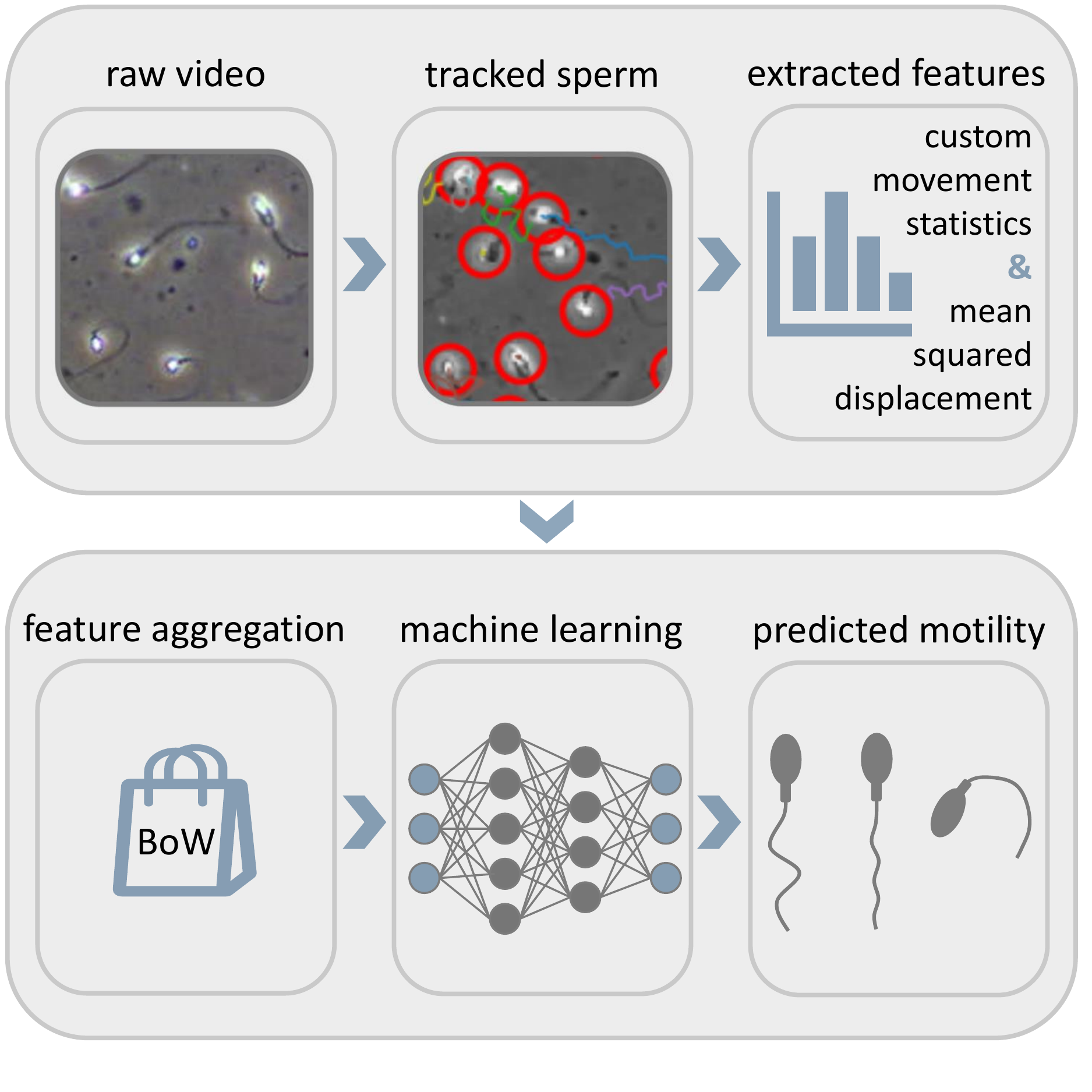}
\end{graphicalabstract}

% \begin{highlights}
% \end{highlights}

\begin{keyword}
computer vision \sep deep learning \sep bag-of-words \sep reproductive health \sep tracking \sep semen motility \sep sperm
% %% MSC codes here, in the form: \MSC code \sep code
% %% or \MSC[2008] code \sep code (2000 is the default)
\end{keyword}

\end{frontmatter}

\section*{SUMMARY}
In this paper, human semen samples from the visem dataset collected by the Simula Research Laboratory are automatically assessed with machine learning methods for their quality in respect to sperm motility. Several regression models are trained to automatically predict the percentage ($0$ to $100$) of progressive, non-progressive, and immotile spermatozoa in a given sample. The video samples are adopted for three different feature extraction methods, in particular custom movement statistics, displacement features, and motility specific statistics have been utilised. Furthermore, four machine learning models, including linear Support Vector Regressor (SVR), Multilayer Perceptron (MLP), Convolutional Neural Network (CNN), and Recurrent Neural Network (RNN), have been trained on the extracted features for the task of automatic motility prediction. Best results for predicting motility are achieved by using the Crocker-Grier algorithm to track sperm cells in an unsupervised way and extracting individual mean squared displacement features for each detected track. These features are then aggregated into a histogram representation applying a Bag-of-Words approach. Finally, a linear SVR is trained on this feature representation. Compared to the best submission of the Medico Multimedia for Medicine challenge, which used the same dataset and splits, the \acf{MAE} could be reduced from $8.83$ to $7.31$. For the sake of reproducibility, we provide the source code for our experiments on GitHub~\footnote{Code available at: \href{https://github.com/EIHW/motilitAI}{https://github.com/EIHW/motilitAI}}.

\section{Introduction}
\label{sec:introduction}
eHealth or ``the use of information and communications technology in support of health and health-related fields''~\cite{world2017global} has been a prioritised item on the agenda of the \ac{WHO} since 2005~\cite{world2005resolution}. From then until 2016, the percentage of \ac{WHO} member states who have a national eHealth policy in place has risen to $58\,\%$~\cite{world2016atlas}.
eHealth is further considered an important factor for improving both the quality and availability of affordable health care, moving countries closer towards achieving universal health coverage~\cite{wha5833sustainable}. 

One such issue can, for example, be found with fertility related problems~\cite{yee2013vivo}. Across the globe, approximately $8-12\,\%$ of couples are affected by infertility~\cite{stephen1998updated, kumar2015trends} which is defined as the inability to achieve clinical pregnancy after 12 or more months of regular unprotected sexual intercourse~\cite{zegers2009international, practice2008definitions}. The issue can be a result of both male and female factor infertility~\cite{kumar2015trends}. In males, infertility is often related to deficiencies in semen quality measured by characteristics and reference values defined by the \ac{WHO}~\cite{cooper2010world}. The attributes most strongly associated with fertility can be found in the concentration, motility and morphology of sperm~\cite{kumar2015trends}. The analysis of these characteristics can serve as a valuable baseline for diagnosis and treatment of patients but requires either specialised, expensive medical equipment, or manual annotation by trained medical staff~\cite{david1981kinematics, mortimer2015future}. In this respect, machine learning approaches that use video recordings of semen samples to detect morphology and motility of the included spermatozoa could assist physicians in their work. To work towards this goal, the \emph{visem} dataset~\cite{visem2019} collected and released by the \emph{Simula Research Laboratory} contains microscopic recordings of semen samples which are additionally annotated with regards to the mentioned characteristics of spermatozoa quality. 

For this paper, a novel combination of unsupervised tracking, feature extraction and quantisation methods, and machine learning models are investigated to perform automatic analysis of the motility of recorded spermatozoa cells. Motility means observing the speed and way of movement of sperm, \ie, if they travel on a straight path or in a circle. Furthermore, prior to extracting features, the data from the \emph{visem} dataset is preprocessed in order to minimise negative impact that might come from blurred camera settings and numerous cuts within each video. The effectiveness of the applied feature extraction methods and machine learning models are compared to the approaches provided by the data organisers and state-of-the-art deep learning-based methodologies from various research groups. These contributions have been submitted to the \emph{Medico: Multimedia for Medicine} sub-challenge~\cite{hicks2019medico} that was part of the 2019 edition of the MediaEval challenge.

The remainder of this paper is structured as follows. The proceeding section reviews the related work on \ac{CASA}, and more specifically on automated prediction of sperm motility. In~\Cref{sec:dataset}, we describe the dataset. \Cref{sec:expset} follows with the illustration of our approach, including preprocessing, particle tracking algorithms, feature extraction and machine learning models. All accomplished results are listed in~\Cref{sec:results} and discussed in~\Cref{sec:discussion}. Finally, we give a conclusion and suggestions for future work in~\Cref{sec:conclusion}.

\section{Related Work}\label{sec:related-work}
Sperm motility and morphology characteristics have been defined in the official \ac{WHO} lab manual~\cite{world1999laboratory}. Here, morphology is measured by defects occurring in three distinct parts of a spermatozoon. Firstly, the head of a sperm cell might show deviations from the norm, \eg, it might be tapered, amorphous or pyriform. Moreover, neck and mid-pieces of cells can be connected to the head in a bent or asymmetrical fashion, or be unusually thin or thick. The motility of sperm cells can further be analysed by classifying them according to multiple categories~\cite{bjorndahl2010practical}. Spermatozoa can either be immotile or motile, where for motile sperm, further categorisation can be applied. A particular cell is motile if its tail is beating. The beating of the tail alone, however, does not translate to effective movement. Therefore, motile sperm cells are additionally grouped according to the progressivity of their movement. Non-progressive spermatozoa beat their tails without any net gain in space, whereas progressive cells do gain space in the process~\cite{bjorndahl2010practical}. 

Traditionally, these characteristics had to be assessed manually by trained clinical staff~\cite{mortimer2015future, david1981kinematics}, but advancements in computational hardware led to the introductions of \ac{CASA}~\cite{mortimer1990objective}. \ac{CASA} works well for many non-human species~\cite{van2009effect, lueders2012improved}, but has traditionally struggled with the accurate assessment of male fertility characteristics from microscopic video recordings of human sperm cells~\cite{bjorndahl2010practical}. This discrepancy is caused both by biological as well as technical limitations~\cite{mortimer1994practical,mortimer1998value, mortimer1995workshop}. First of all, from a biological perspective, human sperm has many characteristics that are detrimental to automatic analysis, such as high amounts of debris particles, generally lower sperm motility and concentration, and many dead spermatozoa which are often also clumped together~\cite{mortimer2015future}. As a consequence, while progressive movement can be detected quite accurately, non-progressive motile sperm cells are very hard to automatically differentiate from drifting debris or dead spermatozoa~\cite{mortimer1995workshop}. Furthermore, the clumping of alive cells with debris or dead spermatozoa can negatively affect the automatic tracking, leading to missing and interrupted tracks~\cite{mortimer2015future}. Morphology is especially hard to assess by commercial \ac{CASA} systems, as accurate analysis is only possible for sperm heads~\cite{mortimer1998value}. Especially for motility, the \ac{CASA} systems base their analysis on computing various kinematic statistics about each sperm track and then using those for determining progressive and non-progressive motility based on agreed upon rules and thresholds~\cite{mortimer2015future}. Therefore, advancements made for these systems are mainly aimed at mitigating the problems and limitations that arise from the general quality of human sperm, such as eliminating drift, recovering sperm tracks through collision or detecting cells that are clumped together~\cite{mortimer2015future}. Urbano~et~al.~\cite{urbano2016automatic}, for example, implemented a robust multi-target sperm tracking algorithm that is able to effectively deal with collisions based on the joint probabilistic data association filter~\cite{bar2009probabilistic}. Apart from the video based \ac{CASA} systems, signal processing based machines, such as the SQA-V Gold Semen Analyzer\footnote{SQA-Vision -- The Ultimate Automated Semen Analysis Solution for Hospitals, Reproductive Centers, Free Standing Labs, and Cryobanks, available at \hyperlink{http://mes-global.com/analyzers}{http://mes-global.com/analyzers}} exist that provide more accurate results but are expensive, prohibiting their use in developing countries. Many \ac{CASA} systems used in research and medical applications are closed-source, proprietary software or integrated hardware-software solutions. However, recently, developments towards the introduction of open-source alternatives into the field have been made, \eg, with openCASA~\cite{10.1371/journal.pcbi.1006691}. Furthermore, applications which solve individual parts of the automatic sperm analysis task can be found with particle tracking software, such as \emph{Trackpy}~\cite{dan_allan_2019_3492186}, or motility analysis toolkits for inference of cell state~\cite{heteromotility}.

The advancements in the field of machine learning, especially \ac{DL} for image analysis, also made an impact on the field, leading to new possibilities for micro cinematographic approaches. \acp{CNN}, for example, have proven their power for a wide range of image analysis tasks~\cite{goodfellow2016deep,chen2014convolutional,donahue2014decaf,zhou2014learning,krizhevsky2012imagenet,oquab2014learning,simonyan2014very,szegedy2015going}. On the publicly available \ac{MHSMA} dataset derived from \ac{HSMA-DS}~\cite{ghasemian2015efficient}, these networks have been applied by Javadi~\etal~\cite{javadi2019novel} to perform automatic morphology classification of spermatozoa images and achieved high detection accuracy for head, vacuole, and tail defect. 
\begin{figure*}[t]
\centering
\begin{subfigure}{.3\textwidth}
\animategraphics[loop,autoplay,width=\columnwidth]{10}{figures/not-tracked/sp-}{0}{39}
\caption{}
\label{subfig:not-tracked-sperm}
\end{subfigure}
\hspace{1cm}
\begin{subfigure}{.3\textwidth}
\animategraphics[loop,autoplay,width=\columnwidth]{10}{figures/tracked/sp-}{001}{039}
\caption{}
\label{subfig:tracked-sperm}
\end{subfigure}
\caption{Microscopic recording of a sperm sample contained in the visem dataset (\Cref{subfig:not-tracked-sperm}). \Cref{subfig:tracked-sperm} depicts spermatazoa tracks as detected by the techniques utilised in this work. \textit{Note: If the pdf file of our manuscript is opened with Acrobat Reader, an animated version (15\,\acs{fps}) of this figure will be played automatically.}}
\label{fig:mel-spectrograms}
\end{figure*}
Dewan~et~al.~\cite{dewan2018estimation} further integrated \ac{CNN} classifiers into their automatic sperm analysis framework as a means to filter out tracks belonging to non sperm particles, such as debris. Applying transfer learning from an ImageNet pre-trained VGG16 network has also been applied to the classification of sperm morphology from individual annotated images~\cite{riordon2019deep}.

In 2019, the Medico Multimedia for Medicine challenge~\cite{hicks2019medico} presented researchers with the opportunity to develop automatic analysis systems for the assessment of human semen quality. The challenge dataset, \emph{visem}, contains 85 video recordings of semen samples which are annotated with regards to morphology and motility of the recorded sperm cells on a per-video basis. 
While there are only a handful of challenge submissions~\cite{hicks2019p56,thambawita2019p62,thambawita2019p61}, they all made use of current deep learning approaches and showed that video based analysis can provide insight into important characteristics of spermatozoa health. For the task of motility prediction, their \ac{CNN} based models could improve significantly over both a ZeroR baseline as well as models based on traditional image features and regression algorithms~\cite{hicks2019p56,hicks2019baseline}.

More related to the methodology applied in this paper, a feature representation of textual documents from the field of Natural Language Processing, namely the Bag-of-Words model, has recently been applied to other domains. One such example can be found in~\cite{amiriparian2018bag,amiriparian2019deep}, where deep feature vectors are aggregated and quantised in an unsupervised fashion to form noise-robust feature representations for a number of audio analysis tasks. Similarly, Amiriparian \etal~\cite{amiriparian2017sentiment} applied Bags-of-Deep-Features for the task of video sentiment analysis. In this work, a similar model is employed to generate feature representations for entire sperm samples from individual per-track movement statistics.

\section{Dataset}\label{sec:dataset}
The data used for the experiments comes from the so-called \emph{visem} \emph{dataset}~\cite{visem2019} collected by the \emph{Simula Research Laboratory}\footnote{visem dataset available at: https://datasets.simula.no/visem/}. This dataset consists of 85 videos of live spermatozoa from men aged 18 years or older. Each video has a resolution of $640 \times 480$ pixels, captured with $400\times$ magnification using an Olympus CX31 microscope and runs at 50 frames-per-second. The name of each video file is composed of the patient ID, the date of recording, a short optional description, and the code of the assessing person (\eg~\textsc{1\_09.09.02\_SSW}). Each sample is annotated with both motility and morphology characteristics. For motility, the percentages ($0$ to $100$) of progressive, non-progressive, and immotile particles are given. These values form the ground truth for the experiments conducted in this paper.

% Moreover, the dataset includes six \emph{csv}-files, one mapping participants' semen samples to the video recordings. 
Further, the dataset includes the results of a standard semen analysis and a set of sperm characteristics, \ie, the level of sex hormones measured in the blood of the participants, levels of fatty acids in spermatozoa or of the phospholipids (measured from the blood). Besides, general, anonymised study participant-related data such as age, abstinence time and Body Mass Index (BMI) is given by the sixth \emph{csv}-file. Additionally, \ac{WHO} analysis data, \eg, ground truth for sperm quality assessment could be accessed. The Medico Multimedia for Medicine challenge's provided subject independent cross-validation folds~\cite{hicks2019medico} were adopted for our experiments.

\section{Approach and Experimental Settings}\label{sec:expset}
\begin{figure*}[!t]
\centerline{\includegraphics[width=\textwidth]{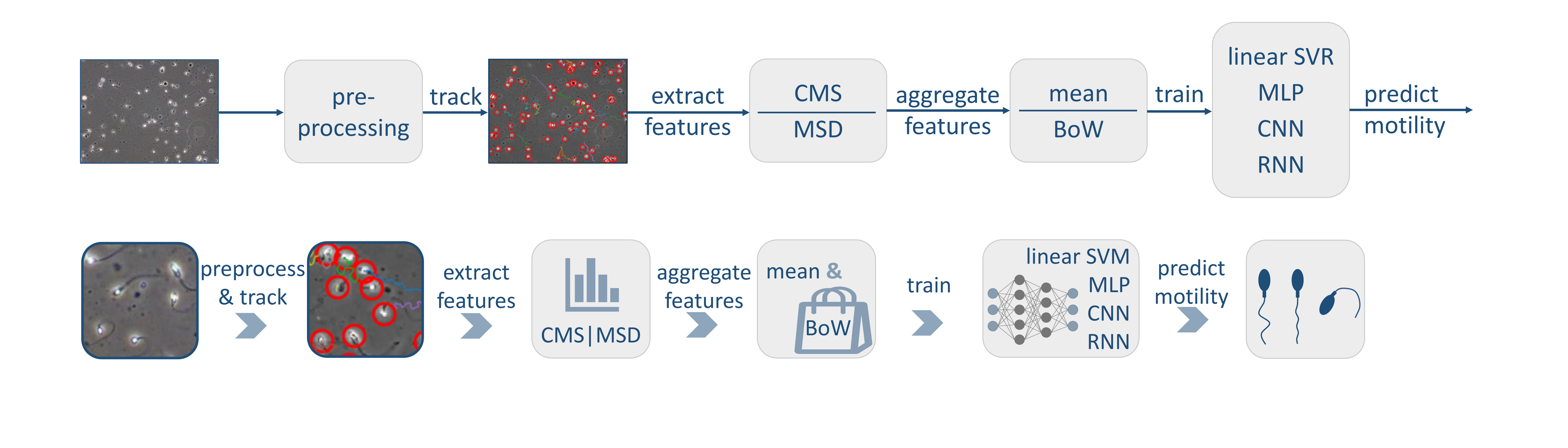}}
\caption{Our proposed framework for motility predictions consists of the following steps. First, preprocessing is applied to the videos after which the spermatozoa are tracked. From these tracks, features are extracted in the form of \emph{custom movement statistics (cms}) and \emph{\acf{MSD}}. Finally, we aggregate those features and use them to train different networks for motility prediction.}
\label{fig:methodology}
\end{figure*}

The different aspects of the overall approach of this paper is depicted in~\Cref{fig:methodology}.
The videos from the dataset are preprocessed and subsequently tracking is applied on them in order to extract features. These features, \ie, \ac{MSD} and movement statistics, are aggregated using \ac{BoW} and, in the case of \ac{MSD}, their mean values. Afterwards, different models, \ie, a linear \ac{SVR}, \ac{MLP} regressor, \ac{CNN} and \ac{LSTM} network, are trained on those features to predict motility. We systematically evaluate all combinations of feature extraction and machine learning models, as far as applicable, \ie, the \ac{BoW} features that form an aggregated, sparse represenation of entire video-samples are not combined with \acp{RNN} or \acp{CNN}.
\subsection{Tracking}\label{ssec:tracking}
In order to achieve tracking of the spermatozoa, two different approaches are pursued. On the one hand, sparse optical flow with the \emph{Lucas-Kanade} algorithm is applied for this purpose, see~\Cref{ssec:tracking-lk}. On the other hand, the \emph{Crocker-Grier} algorithm that is used in the so-called \emph{Trackpy} tool is a second method for tracking spermatozoa, as can be seen in~\Cref{ssec:tracking-cg}.

\subsubsection{Sparse Optical Flow with Lucas-Kanade Algorithm}\label{ssec:tracking-lk}
The Lucas-Kanade method falls into the latter category as a differential approach for estimating sparse optical flow~\cite{lucas1981iterative}. A basic assumption made for computing optical flow is that the brightness of the image is constant across all recorded frames, \ie, pixel intensities are merely translated according to their respective velocities between consecutive video images~\cite{fleet2006optical}. While this assumption rarely holds for real world video sequences, it nevertheless works well in practice for the purpose of estimating optical flow~\cite{fleet2006optical}. The Lucas-Kanade method introduces the additional constraint that the optical flow is constant for any small subspace of the image. Together with Tomasi~\cite{tomasi1991detection}, Kanade improved on this tracking algorithm by detecting good image patches from the eigenvalues of the gradient matrix based on certain thresholds. Shi and Tomasi finally also introduced a method of filtering out bad features, by comparing affine compensated tracked image patches between non-consecutive frames, the assumption being that translation should be enough to account for dissimilarities in image patches along a detected track~\cite{shi1994good}.

Implementations of all of the components that are used in this tracking algorithm are available in the open-source computer vision library OpenCV\footnote{Lucas-Kanade Tracker: https://github.com/opencv/opencv/blob/master/ samples/python/lk\_track.py}~\cite{opencv_library}. 
In order to achieve better results in detecting sperm particles and their positions over time, different values for the feature detection hyperparameters \emph{maxCorners}, \emph{minDistance}, and \emph{blockSize} are optimised to smaller values of $100$, $10$, and $10$, respectively.

However, besides identifying suitable parameter values for tracking the sperm cells, it is necessary to extract information about the sperm's position over time to be able to compute different statistics, such as a certain sperm's speed over a specific time interval. For this purpose, different information on every tracked sperm particle had to be computed and stored. This data includes the number of the first and the last frame of a sperm's track, the position of the sperm in every frame of the track, and the distance the sperm has moved in total. With information stored about all sperm particles in a certain video, features describing a particular spermatozoon's movement can be extracted. The particularities of these features will be discussed in~\Cref{sssec:custom-movement-statistics}.

\subsubsection{Crocker-Grier Algorithm}\label{ssec:tracking-cg}
The second tracking method employed for tracking sperm particles in the videos of the \emph{visem} dataset comes in the form of the Crocker-Grier algorithm for microscopic particle tracking and analysis in colloidal studies~\cite{crocker1996methods}.
This approach's target application is therefore more closely related to the task of spermatozoa tracking from video recordings. The algorithm can track colloidal spheres -- Gaussian-like blobs of a certain total brightness -- across frames of microscopic video recordings of particles, and consists of a number of distinct consecutive steps. First of all, geometric distortion, non-uniform contrast, and noise are alleviated by spatial warping, background removal, and filtering, respectively~\cite{crocker1996methods, jain1989fundamentals, pratt2013introduction}.
After these preprocessing steps, candidate particle centroids can be detected by finding local brightness maxima~\cite{crocker1996methods}, \eg, computed by grayscale dilation~\cite{jain1989fundamentals}. These maxima are further filtered by considering only those in the upper 30th percentile of brightness~\cite{crocker1996methods}, and refined according to the brightness weighted centroids in their immediate vicinity. Afterwards, particle positions can be linked probabilistically considering the dynamics of Brownian particle diffusion~\cite{crocker1996methods}, $P(\delta_i|t)$. 
At this stage, tracks can also be interrupted or terminated if, for example, particles leave the video frame. However, past locations are kept in memory so that relinking is possible, should the particle reappear.

The open-source python library \emph{Trackpy}\footnote{Trackpy tool: https://github.com/soft-matter/trackpy}~\cite{dan_allan_2019_3492186} provides an implementation of this algorithm and further tools for processing and extracting features from particle tracks. Parameters regarding the location and linking of particles into trajectories had to be adjusted in order to improve the tracking accuracy. To reduce the hyper-parameter space of our experimental pipeline, we chose to find these parameters manually by qualitative analysis of a few samples of the \emph{visem} dataset. First, an estimate of $11$ pixels for the size and a minimum mass of $900$ spermatozoa heads is found to lead to accurate detection of sperm cells. For linking the locations, a maximum travel distance of $5$ pixels per frame resulted in consistent, uninterrupted tracking. Furthermore, a maximum of $3$ frame-positions are kept in memory for cells that disappeared. Some of the detected trajectories ($<25$ frames) are too short for analysis and are therefore filtered out. As some videos contain camera drift, \emph{Trackpy}'s built-in drift subtraction is applied. \Cref{fig:methodology} shows a sample frame annotated with sperm locations and respective trajectories as detected by the tool. The drift-subtracted and filtered tracks for each video scene then served as basis for feature extraction, as will be explained in~\Cref{ssec:features}. 

\subsection{Feature Extraction}\label{ssec:features}
For the task of predicting motility statistics for input semen samples, features are extracted based on the spermatozoa tracks obtained with the methods described in~\Cref{ssec:tracking}. Three feature descriptors are considered in the experiments. Custom movement statistics are computed from the tracks generated with the basic \emph{Lucas-Kanade tracker}, mean squared displacement vectors are extracted directly with \emph{Trackpy}, and finally, a range of more involved and computationally heavier particle motility statistics are created.

\subsubsection{Custom Movement Statistics}\label{sssec:custom-movement-statistics}
The first feature representation chosen for performing motility prediction on the dataset is constructed by computing a set of statistics from the tracks detected with the adapted \emph{Lucas-Kanade tracker}. Based on the nature of the task at hand for which it is important to differentiate between progressive and non-progressive movement of sperm cells, both the total amount of movement by spermatozoa in particular time frames as well as the actual distances covered by them are of interest. The first aspect can be calculated for a specific window by accumulating the number of pixels a particular cell moved between each consecutive video frame while the second metric looks at the Euclidean distance between the positions of the cell at the start and end of the time window. These calculations can then be carried out for sliding windows of different sizes, and statistical functionals can be applied to their results. This leads to feature vectors of fixed length for each sperm track found by the Lucas-Kanade tracking algorithm. Specifically, these window sizes are used (measured in number of frames): $5$, $10$, $20$, $50$, $80$, $100$, $150$, $200$, $250$, $300$, $400$, $500$, $750$, and $1\,000$. After computing both metrics as described above for the whole sample by sliding each of the windows over a particular track with a hop size of one frame, mean, maximum, and minimum functionals are applied to the resulting series of motility calculations. Two additional features are computed as the total distance covered by a single sperm cell during the whole video sample and its average speed in pixels moved per frame. In total, the approach leads to numerical feature vectors of size $14\times2\times3 + 2 = 86$ for each detected sperm track. Before being applicable to the task of motility analysis on a per-sample-basis, these vectors can be further processed and aggregated per video clip. Here, two possibilities are explored. First, feature vectors of a single video sample are reduced by their feature-wise mean. Secondly, a \ac{BoW} approach is applied to the vectors that both quantises and summarises them in an unsupervised manner.

\subsubsection{Displacement Features}\label{sssec:features-displacement}
A common statistical measure that is employed to characterise the random movement of particles can be found with the \ac{MSD}~\cite{frenkel2001understanding}. It can be used to describe the explorative behaviour of particles in a system, \ie, if movement is restricted to diffusion or affected by some sort of force. The displacement of a single particle $j$ is defined as the distance it travelled in a particular time frame of duration $l$ (lag-time) $t_{i}$ to $t_{i+l}$ measured as the square of the Euclidean distance between its positions at the start ($x_j(t_{i})$) and end ($x_j(t_{i+l})$) of the frame. For a set of $N$ particles, the \emph{ensemble mean} displacement for a specific time interval can then be computed as:
\begin{equation}
    MSD = \left < |x(t_{i+l}) - x(t_{i}) |^{2}) \right > = \frac{1}{N} \sum_{j=1} ^{N} |x_j(t_{i+l}) - x_j(t_{i}) |^{2}. 
\end{equation}
When observing a longer period of time ($T_0$ to $T_1$), an average of \ac{MSD} can further be computed from sliding windows of particular lag times over the whole segment. This can be done for each individual particle (\ac{imsd}) or again as an average for all of the particles (\ac{emsd}). Finally, computing these displacement values for a range of different lag times can capture more detailed information about particle movement. For the application of automated sperm motility analysis, mean squared displacement of spermatozoa in a given sample for different sized time windows could give insight into the amount of progressive and non-progressive motility. Given enough time, a progressive sperm cell would travel across a larger distance, whereas a sperm that is merely moving in place would display the same amount of displacement for both short and long time frames. \emph{Trackpy} provides interfaces to compute both \emph{\ac{imsd}} and \emph{\ac{emsd}} for a range of increasing time frames. Specifically, it considers lag-times up to a user definable maximum that are increased in frame wise step sizes, \ie, in the case of the \emph{visem} dataset that is recorded at 50\,fps, the consecutive window sizes grow by $20\,ms$, each. When considering a maximum lag-time of ten seconds for example, 500 mean squared displacement values are computed from the sperm tracks. As \emph{\ac{emsd}} is computed as an aggregated measurement for all sperm cells of a given sample in a particular time frame, it can be directly used as input for the machine learning algorithms described in~\Cref{ssec:classification-models}. Also, \emph{\ac{imsd}} feature vectors which are extracted on a per-track-basis, can be further quantised and aggregated using the Bag-of-Words framework described in~\Cref{ssec:BoW} to form a clip level representation. In this article, three different combinations of window and hop sizes are considered for the extraction of \emph{\ac{emsd}} feature vectors: a window size of 2\,seconds with a 1\,second hop and 10\,seconds windows with either 1\,second or 5\,seconds hops. Based on the motility prediction performance achieved using the different \emph{\ac{emsd}} feature configurations, hop and window sizes for \emph{\ac{imsd}} prediction are chosen.

\subsection{Bag-of-Words}\label{ssec:BoW}
The use of unsupervised tracking algorithms allows the extraction of useful features on a more granular, per-spermatozoon basis. As the sperm cell count varies heavily between the different samples in the \emph{visem} dataset and annotations are further only available on a per-sample level, some sort of feature aggregation mechanism has to be implemented to leverage per-cell information. 
\begin{table}[h]
\centering
\resizebox{\columnwidth}{!}{
\begin{tabular}{llll}
    \textbf{hyperparameter} & \textbf{\ac{MLP}} & \textbf{\ac{CNN}} & \textbf{\acs{RNN}} \\
    \toprule
    batch size & $16$, $32$, $64$ & $16$, $32$, $64$ & $16$, $32$, $64$ \\
    dropout & $0.2$, $0.4$ & $0.2$, $0.4$ & $0.2$, $0.4$ \\
    kernel regulariser & $10^{-4}$, $10^{-3}$, $10^{-2}$ & $10^{-4}$, $10^{-3}$, $10^{-2}$ & $10^{-4}$, $10^{-3}$, $10^{-2}$ \\
    activation dense & \acs{ELU}, \acs{ReLU} & \acs{ELU}, \acs{ReLU} & \acs{ELU}, \acs{ReLU} \\
    number of layers & $2$, $4$, $8$ & $2$, $4$, $8$ & $2$, $4$ \\
    learning rate & $10^{-4}$, $10^{-3}$, $10^{-2}$ & $10^{-3}$, $10^{-2}$ & $10^{-4}$, $10^{-3}$ \\
    no.\ of units/filters & $256$, $512$, $1\,024$ & $32$, $64$ & $64$, $128$, $256$ \\
    cell type & -- & -- & \acs{GRU}, \ac{LSTM} \\
    recurrent dropout & -- & -- & $0$, $0.2$, $0.4$ \\
    bidirectional & -- & -- & true, false \\
    \bottomrule
\end{tabular}
}
\caption{All hyperparameters and their values that are optimised for the different machine learning models.}
\label{tbl:hyperparameters}
\end{table}
In~\Cref{sssec:features-displacement}, regarding the mean displacement of all spermatozoa during a given time frame of a specific recording has been introduced as a first, baseline method for this problem. However, simply averaging the displacement of all cells might lead to the loss of more granular information. For this reason, a histogram representation based on the famous \ac{BoW} model extended to be used with arbitrary numerical input features will be employed. For the experiments, the input feature vectors belonging to individual sperm cell tracks are first standardised to zero mean and unit variance before a random subset is chosen to form a codebook. Afterwards, a fixed number of the top nearest vectors from the codebook is computed for each input feature vector. Aggregated over all sperm tracks belonging to a given sample recording, the counts of these assigned vectors form a histogram representation which is further processed by \ac{tf-idf}. Furthermore, the number of codebook vectors $N$ and assigned vectors $a$ are optimised by evaluating all combinations of $N \in [2\,500, 5\,000, 10\,000]$ and $a \in [1, 10, 50, 100, 200, 500]$ on the given data using different machine learning models.

\subsection{Classification Models}\label{ssec:classification-models}
The features that are described in~\Cref{ssec:features} are used as input for various Machine Learning approaches. The extraction methods described above lead to variable numbers of feature vectors for each original video sample, \eg, displacement vectors are extracted for overlapping windows. In order to enable comparisons between all implemented approaches and the methods applied by the participants of the Medico challenge, the predictions of each model are mean aggregated on a per sample basis, \ie, each model produces a single prediction for each of the $85$ patients contained in the \emph{visem} dataset. The models are outlined in the following.

\subsubsection{Linear Support Vector Regressor}\label{ssec:SVR-model}
The first method to predict the motility of spermatozoa is a linear \ac{SVR}. Here, different scaling options, \ie, \emph{StandardScaler}, \emph{MinMaxScaler}, and no scaler, are tested. Five distinct complexity values $c$ equally distributed between $10^{-1}$ and $10^{3}$ are evaluated. The best value for this is found by the \ac{MAE} obtained in an internal 5-fold cross-validation on each fold's training data.

\subsubsection{Multilayer Perceptron}\label{ssec:MLP-model}
The architecture of the \ac{MLP} model contains multiple fully connected layers with batch normalisation applied before the activation function. The model is trained with the \emph{Adam} optimiser in order to minimise the \ac{MSE} and an additional L2 weight regularisation term. \acf{ELU} and \acf{ReLU} are evaluated as choices for the activation functions of the layers. 
A random search is performed over different parameters, including learning rate, number of layers and units per layer, batch size, and dropout that are listed in~\Cref{tbl:hyperparameters}. The best parameters are determined by the \ac{MAE} achieved on the random 20\,\% validation splits of each fold's training data.

\subsubsection{Convolutional Neural Network}\label{ssec:CNN-model}
\begin{figure}[!t]
\centerline{\includegraphics[width=\columnwidth]{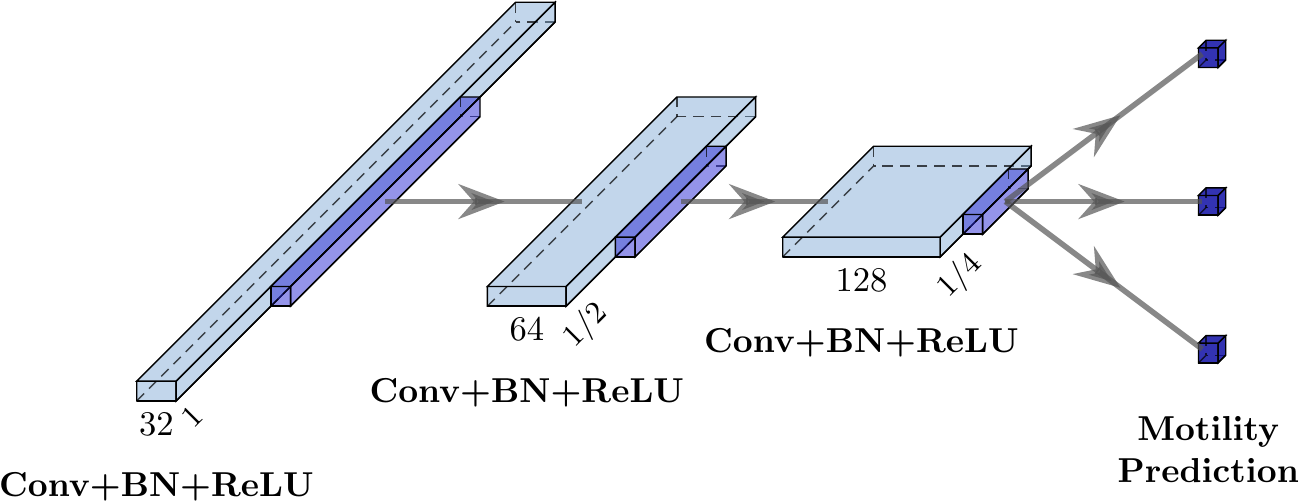}}
\caption{This figure shows the architecture of the model used for the CNN. Three similar blocks of layers with an increasing number of output filters are stacked consecutively. Finally, the output of the last block is fed into a fully connected layer with output neurons for each of the predicted motility characteristics \emph{percentage of immotile sperm}, \emph{percentage of progressive motility}, and \emph{percentage of non-progressive motility}.}
\label{fig:CNN-model-architecture}
\end{figure}

Another method used for the prediction of motility of spermatozoa is a 1-dimensional CNN. Its model architecture is constructed by multiple convolutional blocks which are stacked on top of each other. Each convolutional block consists of the following parts. First, a 1-dimensional convolutional layer with a kernel size of $3$ and stride of $1$ extracts features from the input. Batch normalisation is then applied before the non-linear activation function. Afterwards, the output is max pooled and neurons are randomly dropped out to prevent overfitting. Furthermore, the number of filters in the convolutional layer is doubled for each consecutive block. After the last block, a fully-connected layer with linear activation predicts the three target values for the regression problems. 
\Cref{fig:CNN-model-architecture} depicts an example of such a \ac{CNN} with $32$ filters in the first layer and three convolutional blocks. The model is trained with the \emph{Adam} optimiser to minimise the \ac{MSE} and an additional L2 weight regularisation term. In order to optimise both model architecture and training settings, a random parameter search is performed. Here, the learning rate of the \emph{Adam} optimiser, different functions for the activation, the number of layers, filters and batch size, dropout, and kernel regulariser are adjusted, as can be seen in~\Cref{tbl:hyperparameters}. $50$ different combinations of those parameters are tested and the best one is chosen according to validation \ac{MAE}. The network is trained for an indefinite number of epochs, stopping early if the validation \ac{MAE} has not increased for $100$ epochs.

\begin{table*}[h]
\centering
\resizebox{.8\textwidth}{!}{
\begin{tabular}{lrrrrrrrrrr}
    &&& \multicolumn{2}{c}{\textbf{\ac{SVR}}} & \multicolumn{2}{c}{\textbf{\ac{MLP}}} & \multicolumn{2}{c}{\textbf{CNN}} & \multicolumn{2}{c}{\textbf{\ac{RNN}}} \\
    \textbf{metric} & \textbf{hop} & \textbf{window} & \textbf{val} & \textbf{eval} & \textbf{val} & \textbf{eval} & \textbf{val} & \textbf{eval} & \textbf{val} & \textbf{eval} \\
    \toprule
    \multirow{3}{*}{\textbf{\ac{MAE}}} & $1$\,s & $2$\,s & $11.13$ & $10.91$ & $6.49$ & $11.56$ & $6.29$ & $10.48$ & $6.55$ & $12.97$ \\
    & $1$\,s & $10$\,s & $10.16$ & $8.36$ & \cellcolor{lightgray}\textbf{5.19} & \cellcolor{lightgray}\textbf{8.83} & \textbf{5.03} & \textbf{8.44} & $6.59$ & $8.49$ \\
    & $5$\,s & $10$\,s & \textbf{10.12} & \textbf{8.60} & $5.26$ & $8.13$ & $7.21$ & $8.74$ & \textbf{6.45} & \textbf{8.13} \\
    \midrule
    \multirow{3}{*}{\textbf{\acs{RMSE}}} & $1$\,s & $2$\,s & $14.02$ & $14.30$ & $8.04$ & $15.13$ & $8.21$ & $14.17$ & $10.27$ & $16.17$ \\
    & $1$\,s & $10$\,s & $12.87$ & $11.06$ & \textbf{6.84} & \textbf{11.48} & \cellcolor{lightgray}\textbf{6.55} & \cellcolor{lightgray}\textbf{10.82} & \textbf{9.18} & \textbf{11.41} \\
    & $5$\,s & $10$\,s & \textbf{12.85} & \textbf{11.56} & $6.95$ & $10.56$ & $7.21$ & $11.31$ & $9.49$ & $10.79$ \\
    \bottomrule
\end{tabular}
}
\caption{\acf{MAE} and \acf{RMSE} results of proposed experiments using four machine learning models on \emph{\ac{emsd}} features described in~\Cref{ssec:SVR-results,ssec:MLP-results,ssec:CNN-results,ssec:LSTM-results}. Three different \emph{hop} and \emph{window size} combinations are evaluated for the extraction of the \emph{\ac{emsd}} feature vectors.}
\label{tbl:results-mae}
\end{table*}

\subsubsection{Recurrent Neural Network}\label{ssec:LSTM-model}
The model architecture of the considered \acf{RNN} consists of multiple recurrent layers. Each of those layers contains either a \acf{GRU} or \ac{LSTM} cell. Bidirectional variants where the input is processed both in the forward and backward direction are also tested. Dropout is applied both within (between time steps) and after each recurrent layer. Both \acp{GRU} and \acp{LSTM} use hyperbolic tangent activation (\emph{\ac{tanh}}) for their recurrent layers and \emph{sigmoid} as their gate activation functions. In order to minimise the \ac{MSE} and an additional L2 weight regularisation term, the model is trained with the \emph{RMSProp} optimiser. Parameters such as learning rate, number of layers and recurrent units, and batch size are optimised, as listed in~\Cref{tbl:hyperparameters}.

\section{Results}\label{sec:results}
Motility of spermatozoa is predicted by a linear \ac{SVR} (\Cref{ssec:SVR-results}), an \ac{MLP} (\Cref{ssec:MLP-results}), a \ac{CNN} (\Cref{ssec:CNN-results}), and an \ac{LSTM} (\Cref{ssec:LSTM-results}), where all models are trained on \emph{\ac{emsd}} features extracted with \emph{Trackpy}. Moreover, motility prediction is achieved by \ac{BoW} with a linear \ac{SVR} (\Cref{ssec:BoW-SVR-results}) and with an \ac{MLP} regressor (\Cref{ssec:BoW-MLP-results}), both trained on \emph{\ac{imsd}} features extracted with the help of \emph{Trackpy}, and on features created by computations on a set of statistics from tracks detected with the customised \emph{Lucas-Kanade tracker}. We do did not train any \ac{CNN} or \ac{RNN} models on the \acp{BoW}, as they are sparse quantisations of entire video samples, thus containing neither structural nor temporal information that could be exploited by those types of neural networks. 

\ac{MAE} and \acf{RMSE} results for both validation and evaluation are outlined in~\Cref{tbl:results-mae,tbl:results-BoW}. However, for purposes of readability, in the text, we mainly remark on \ac{MAE} results achieved on evaluation using the same parameters found for the best results on validation.

\subsection{Linear Support Vector Regressor}\label{ssec:SVR-results}
The first set of results comes from training a linear \ac{SVR} on the \emph{\ac{emsd}} feature vectors extracted with \emph{Trackpy}. As described in~\Cref{sssec:features-displacement}, three different combinations of window and hop size are evaluated for feature extraction. Training and optimisation of the regressor is further done as outlined in~\Cref{ssec:SVR-model}. 
\Cref{tbl:results-mae} shows both \acp{MAE} and \acp{RMSE} achieved during validation and evaluation using three-fold cross-validation. It is apparent from both the validation and evaluation results, that choosing only a small window size of two seconds for computing the displacement statistics leads to feature representations that lack useful information for predicting motility characteristics of the sperm cells for each sample. For the configurations using a larger window size of ten seconds and a larger hop size of five seconds leads to slightly better results during validation but decreased evaluation performance. Considering the best validation result, a minimum \ac{MAE} of $8.60$ is obtained on evaluation with the \ac{SVR} trained on \emph{\ac{emsd}} features. Measured against the state-of-the-art, this result shows a relative improvement of 2.6\,\%~\cite{thambawita2019p61}.

\subsection{Multilayer Perceptron}\label{ssec:MLP-results}
Secondly, an \ac{MLP} is trained on the \emph{\ac{emsd}} feature vectors that have been extracted with \emph{Trackpy}. Again, following the procedure described in~\Cref{sssec:features-displacement}, features are extracted by three combinations of window and hop size. In~\Cref{ssec:MLP-model}, it is shown how the network is trained and optimised. Best results are achieved with a learning rate of $10^{-2}$, a batch size of $16$, and a dropout of $0.2$. For a window size of two seconds and one second hop, best results are obtained with the \ac{ReLU} activation function, $8$ layers, a factor of $10^{-4}$ for the L2 weight regularisation and $1\,024$ units on each layer. The model trained on features extracted with a window size of ten seconds and one and five seconds hop are performing best for choosing \ac{ELU} as activation function, $4$ layers, $10^{-3}$ as factor for the L2 weight regularisation, and $512$ units on each layer. In~\Cref{tbl:results-mae}, an overview of \ac{MAE} and \ac{RMSE} results coming from validation and evaluation with three-fold cross-validation is given. Same as with the \ac{SVR} (see~\Cref{ssec:SVR-results}), choosing a window size of two seconds for the computation of displacement statistics performs the worst. For choosing a larger window size of ten seconds, validation performance is somewhat better for applying a hop size of one second than five seconds. However, the larger hop size of five seconds is performing slightly better on evaluation. The minimum \ac{MAE} value of $8.83$ is achieved by the \ac{MLP} trained on displacement features, which is as good as the findings of \ac{SOTA}~\cite{thambawita2019p61}.

\begin{table*}[h!]
\centering
\begin{tabular}{rrrrrrrrrr}
    && \multicolumn{2}{c}{\textbf{\ac{SVR} + cms}} & \multicolumn{2}{c}{\textbf{\ac{SVR} + msd}} & \multicolumn{2}{c}{\textbf{\ac{MLP} + cms}} & \multicolumn{2}{c}{\textbf{\ac{MLP} + msd}} \\
    \textbf{codebook size} & \textbf{assigned vectors}  & \textbf{val} & \textbf{eval} & \textbf{val} & \textbf{eval} & \textbf{val} & \textbf{eval} & \textbf{val} & \textbf{eval} \\
    \toprule
    \multirow{6}{*}{2\,500} & 1 & 8.34 & 8.00 & 7.71 & 7.55 & 7.63 & 8.11 & 6.70 & 8.01 \\
    & 10 & 8.31 & 7.91 & 7.85 & 7.73 & 7.37 & 8.37 & 6.19 & 8.74 \\
    & 50 & 8.33 & 8.03 & 8.10 & 8.01 & 7.52 & 8.50 & 6.29 & 7.95 \\
    & 100 & 8.28 & 8.00 & 8.18 & 8.06 & 7.18 & 8.64 & 6.47 & 8.35 \\
    & 200 & 8.23 & 8.05 & 8.26 & 8.09 & 7.15 & 8.39 & 6.41 & 8.16 \\
    & 500 & 8.26 & 8.31 & 8.27 & 8.03 & \textbf{6.75} & \textbf{8.11} & \cellcolor{lightgray}\textbf{5.91} & \cellcolor{lightgray}\textbf{7.85} \\
    \midrule
    \multirow{6}{*}{5\,000} & 1 & 8.42 & 8.26 & 7.87 & 7.81 & 7.99 & 8.29 & 6.70 & 8.04 \\
    & 10 & 8.22 & 7.93 & 7.69 & 7.43 & 7.12 & 8.73 & 6.65 & 8.54 \\
    & 50 & 8.38 & 8.07 & 8.05 & 7.99 & 7.50 & 8.42 & 6.30 & 8.18 \\
    & 100 & 8.34 & 8.03 & 8.11 & 8.03 & 7.39 & 8.40 & 6.14 & 8.54 \\
    & 200 & 8.29 & 8.00 & 8.19 & 8.07 & 7.31 & 8.40 & 6.35 & 8.68 \\
    & 500 & 8.22 & 8.08 & 8.28 & 8.10 & 7.05 & 7.54 & 6.23 & 8.40 \\
    \midrule
    \multirow{6}{*}{10\,000} & 1 & 8.56 & 8.73 & 8.08 & 8.18 & 7.85 & 7.92 & 7.03 & 8.11 \\
    & 10 & \textbf{8.19} & \textbf{7.86} & \cellcolor{lightgray}\textbf{7.56} & \cellcolor{lightgray}\textbf{7.31} & 7.41 & 8.17 & 6.47 & 8.27 \\
    & 50 & 8.40 & 7.98 & 7.92 & 7.86 & 7.41 & 8.17 & 6.27 & 8.07 \\
    & 100 & 8.38 & 8.07 & 8.05 & 7.99 & 7.42 & 7.95 & 6.12 & 8.04 \\
    & 200 & 8.34 & 8.03 & 8.11 & 8.03 & 7.26 & 8.26 & 6.28 & 8.19 \\
    & 500 & 8.27 & 8.03 & 8.23 & 8.09 & 7.13 & 7.67 & 6.34 & 7.91 \\
    \bottomrule
\end{tabular}
\caption{\acf{MAE} results of proposed experiments using a \ac{BoW} with \ac{SVR} and \ac{MLP} on \emph{\acf{cms}} and \emph{\acf{MSD}} features described in~\Cref{ssec:BoW-SVR-results,ssec:BoW-MLP-results}. 18 different \emph{codebook sizes} and number of \emph{assigned vectors} combinations, all with a window size of five seconds are evaluated.}
\label{tbl:results-BoW}
\end{table*}

\subsection{Convolutional Neural Network}\label{ssec:CNN-results}
Features extracted with the help of the \emph{Trackpy} tool are used for a third set of experiments, this time applying a CNN. Here, features are extracted by three combinations of window and hop size, as can be read in~\Cref{sssec:features-displacement}. How the network is trained and optimised is explained in~\Cref{ssec:CNN-model}. Best results are achieved with a learning rate of $10^{-2}$, \ac{ELU} as activation function, and a factor of $10^{-2}$ for the L2 weight regularisation. Training the network on features extracted with a window size of two seconds and one second hop performed best with $4$ layers, starting with $32$ filters for the first layer, a batch size of $64$, and a dropout of $10^{-2}$. Best results with a window size of ten seconds and hop size of one and five seconds are for a number of $8$ layers, starting with $32$ filters for the first layer, a batch size of $32$, and a dropout of $0.4$. As can be observed in~\Cref{tbl:results-mae}, using the \emph{\ac{emsd}} features obtained with one second overlapping two second segments of one video is not quite keeping pace with the best results of previous papers on this kind of experiment. \emph{\ac{emsd}} features with overlapping ten second parts of the videos with a hop size of one and five seconds are more promising. Experiments with one second hop are achieving best results for \ac{MAE} and \ac{RMSE} values on validation and evaluation. Going by the best validation result, the minimum \ac{MAE} is at $8.44$ for training a \ac{CNN} on \emph{\ac{emsd}} features. These results show a relative improvement of $4.4$\,\% against state-of-the-art results by Thambawita \etal~\cite{thambawita2019p61}. Moreover, the \ac{CNN} performs slightly better than the previous models -- \ac{SVR} achieving $8.60$~\ac{MAE}, see~\Cref{ssec:SVR-results}, and \ac{MLP} resulting in $8.83$~\ac{MAE}, see~\Cref{ssec:MLP-results}. As the improvements are only marginal at best, it is questionable if structural dependencies which could be exploited by the CNN can be found in the \emph{\ac{emsd}} feature vectors.

\subsection{Recurrent Neural Network}\label{ssec:LSTM-results}
A fourth set of experiments is done with training a \ac{RNN} on the features extracted with the \emph{Trackpy} tool. As described in~\Cref{sssec:features-displacement}, three different combinations of window and hop size are assessed for feature extraction. Now, contrary to the other experiments, the feature sequences are formed from the vectors extracted from consecutive overlapping windows and the \ac{RNN} uses all of the information in the sequence to make a prediction. The network is trained and optimised according to~\Cref{ssec:LSTM-model}. Best results are obtained with bidirectional \ac{LSTM} cells with a number of $256$ recurrent units on each layer. Further, applying dropout to the activations in the recurrent layers decreases performance in all cases. For training this network on features of two and ten seconds windows and one second hop, the best hyperparameters are a learning rate of $10^{-3}$, a number of $2$ layers, a batch size of $16$, a dropout of $0.4$, a factor of $10^{-2}$ for the L2 weight regularisation. With a window size of ten seconds and five seconds hop best performance is achieved with a learning rate of $10^{-4}$, a number of $4$ layers, a batch size of $64$, a dropout of $0.2$, a factor of $10^{-4}$ for the L2 weight regularisation. Validation and evaluation with three-fold cross-validation scored the \ac{MAE} and \ac{RMSE} values displayed in~\Cref{tbl:results-mae}. Choosing a window size of ten seconds and a hop size of five seconds achieves evaluation results that are slightly better than state-of-the-art results. The minimum \ac{MAE} is at $8.13$~\ac{MAE} for evaluation, a relative improvement of $7.8$\,\% against state-of-the-art results by Thambawita \etalns~\cite{thambawita2019p61}, as well as $3.6$\,\% against previous experiments, for example those using a \ac{CNN} on the same \emph{\ac{emsd}} features, cf.~\Cref{ssec:CNN-results}.  Considering the temporal dependencies within sequences of \emph{\ac{emsd}} vectors therefore seems to improve on regression performance if ever so slightly. 
\begin{table*}[h]
\centering
\resizebox{.8\textwidth}{!}{
\begin{tabular}{ll|rrrr|rr|rr|r}
    && \multicolumn{4}{c|}{\textbf{emsd}} & \multicolumn{2}{c|}{\textbf{\acs{cms}}} & \multicolumn{2}{c|}{\textbf{imsd}} & \\
    && \textbf{\ac{SVR}} & \textbf{\ac{MLP}} & \textbf{CNN} & \textbf{\ac{RNN}} & \textbf{\ac{SVR}} & \textbf{\ac{MLP}} & \textbf{\ac{SVR}} & \textbf{\ac{MLP}} & \textbf{\acs{SOTA}} \\
    \toprule
    \multirow{2}{*}{\textbf{\ac{MAE}}} & \textbf{val} & $10.12$ & $5.19$ & $5.03$ & $6.45$ & $8.19$ & $6.75$ & $7.56$ & $5.91$ & -- \\
    & \textbf{eval} & $8.60$ & $8.83$ & $8.44$ & $8.13$ & $7.86$ & $8.11$ & \textbf{7.31} & $7.85$ & $8.83$ \\
    \midrule
    \multirow{2}{*}{\textbf{\ac{RMSE}}} & \textbf{val} & $12.85$ & $6.84$ & $6.55$ & $9.18$ & $10.16$ & $9.04$ & $9.39$ & $8.74$ & -- \\
    & \textbf{eval} & $11.56$ & $11.48$ & $10.82$ & $11.41$ & $10.38$ & $10.81$ & $9.56$ & $10.49$ & $12.05$ \\
    \bottomrule
\end{tabular}
}
\caption{\acf{MAE} and \acf{RMSE} results of proposed experiments using eight different machine learning models with \ac{SVR}, \ac{MLP}, \ac{CNN}, and \ac{RNN} on \emph{\ac{emsd}}, \emph{\ac{imsd}}, and \emph{\acf{cms}} features described in~\Cref{ssec:SVR-results,ssec:MLP-results,ssec:CNN-results,ssec:LSTM-results,ssec:BoW-SVR-results,ssec:BoW-MLP-results} compared to \acf{SOTA} results~\cite{thambawita2019p61,hicks2019p56}.}
\label{tbl:results}
\end{table*}
Furthermore, the \ac{RNN} experiments enforce the notion that \emph{\ac{emsd}} features computed over longer time intervals contain more information regarding the motility of sperm cells, as even when taking a sequence of shorter frames into account as a whole, results are better with the greater window size.

\subsection{Bag-of-Words with Support Vector Regressor}\label{ssec:BoW-SVR-results}
A possible drawback of the initial experiments with \emph{\ac{emsd}} features might be that they aggregate information about the movement across all spermatozoa in a given sample in a very primitive fashion with mean values. Therefore, the experiments with unsupervised feature quantisation and aggregation via \ac{BoW} of single-spermatozoon based features investigate a more sophisticated approach of analysing a variable number of sperm cells.

Here, the prediction of the motility of spermatozoa is accomplished by generating \acp{BoW} from the features described in~\Cref{ssec:features} that serve as input for training a \ac{SVR}.

\subsubsection{Custom Movement Statistics Features}\label{sssec:results-custom}
In the first set of experiments for predicting motility using a \ac{BoW}, the \ac{BoW} is generated from movement statistics coming from the adapted \emph{Lucas-Kanade tracker}, as discussed in~\Cref{sssec:custom-movement-statistics}. As shown in~\Cref{ssec:BoW,ssec:SVR-model}, for training and optimisation of the model, codebook sizes of $2\,500$, $5\,000$, and $10\,000$, assigning $1, 10, 50, 100, 200$, and $500$ vectors, and complexity values between $10^{-1}$ and $10^{3}$ are considered. 
Best results are detected with a complexity of $10$. Validation and evaluation achieved \ac{MAE} results are shown in~\Cref{tbl:results-BoW}. Choosing any of the investigated combinations of codebook size and the number of assigned vectors leads to evaluation results that are slightly better than state-of-the-art results by Thambawita \etalns~\cite{thambawita2019p61}. The best result on the validation, achieved with a codebook size of $10\,000$ and $10$ assigned vectors is $7.86$~\ac{MAE} on evaluation, a relative improvement of $10.9$\,\% against state-of-the-art results~\cite{thambawita2019p61} and $6.8$\,\% against the best result of previous experiments with a \ac{CNN} cf.~\Cref{ssec:CNN-results}. These results suggest the superiority of the \ac{BoW} approach to simple mean aggregation. Tuning the \ac{BoW} hyperparameters shows that choosing a smaller number of codebook vectors to assign to each input sample leads to improved results. Furthermore, a marginal performance gain can be achieved with larger codebooks.

\subsubsection{Displacement Features}\label{sssec:BoW-SVR-displacement-results}
The same model as in the preceding part, the \ac{BoW} with a linear \ac{SVR}, is now trained on \emph{\ac{imsd}} features extracted with \emph{Trackpy}, described in~\Cref{sssec:features-displacement}. As outlined in~\Cref{ssec:BoW,ssec:SVR-model}, codebook sizes of $2\,500, 5\,000$, and $10\,000$, assigning $1, 10, 50, 100, 200$, and $500$ vectors and complexity values between $10^{-1}$ and $10^{3}$ are tested to extract features in order to train and optimise the model. A complexity of $10^{3}$ showed the best results. In~\Cref{tbl:results-BoW}, the validation and evaluation for \ac{MAE} values are reported. 
Evaluation results for any tested combination of codebook size and number of assigned vectors is better than or at least equally good as all previous experiments in this article and state-of-the-art results. A codebook size of $10\,000$ and $10$ assigned vectors achieves the overall minimum evaluation \ac{MAE} of \textbf{$7.31$}. This is a relative improvement of over $17.2$\,\% compared to the best submission~\cite{thambawita2019p61} and is outperforming the results provided in~\Cref{sssec:results-custom}. 

The \emph{\ac{imsd}} features extracted with \emph{Trackpy} therefore serve as a more powerful basis for feature creation than the custom statistics generated from movement tracks. The observations about codebook sizes and number of assigned vectors also hold for this set of experiments, with larger codebooks and fewer vector assignments leading to the best results.

\subsection{Bag-of-Words with Multilayer Perceptron}\label{ssec:BoW-MLP-results}
Motility prediction of spermatozoa is additionally achieved by training a \ac{BoW} with an \ac{MLP} on features created with both \emph{Trackpy} and calculations coming from the adapted \emph{Lucas-Kanade tracker}.

\subsubsection{Custom Movement Statistics Features}\label{sssec:BoW-MLP-results-lk-track}
Experiments for this model are started by training the \ac{BoW} with an \ac{MLP} regressor on custom movement statistics features created with the help of the tailored \emph{Lucas-Kanade tracker} that has been explained in~\Cref{sssec:custom-movement-statistics}. Codebook sizes of $2\,500, 5\,000$, and $10\,000$ and $1$, $10$, $50$, $100$, $200$, and $500$ assigned vectors are tested for feature extraction, so that the model can be optimised, see~\Cref{ssec:BoW}. For further optimisation, various values for different hyperparameters are assessed as shown in~\Cref{ssec:MLP-model}. Best results are accomplished with a learning rate of $10^{-2}$, a batch size of $64$, a dropout of $0.4$, $2$ layers, and $1\,024$ units per layer. \ac{ReLU} is the best performing activation function and a factor of $10^{-2}$ proved best for the L2 weight regularisation. \ac{MAE} results for validation and evaluation are listed in~\Cref{tbl:results-BoW}.
The best \ac{MAE} results for evaluation of $8.11$~\ac{MAE} are achieved for a codebook size of $2500$ and $500$ assigned vectors.

\subsubsection{Displacement Features}\label{sssec:BoW-MLP-results-displacement}
The same model of a \ac{BoW} with an \ac{MLP} regressor as in the preceding part in this section is additionally trained on displacement features extracted with \emph{Trackpy}, as shown in~\Cref{sssec:features-displacement}. As described in~\Cref{ssec:BoW}, codebook sizes of $2\,500, 5\,000$, and $10\,000$ and $1, 10, 50, 100, 200$, and $500$ assigned vectors are evaluated for feature extraction. This model is further trained and optimised according to~\Cref{ssec:MLP-model}, obtaining best results with a learning rate of $10^{-2}$, a batch size of $16$, a dropout of $0.2$, $4$ layers and $256$ units in each of those layers. Here, \ac{ELU} activation function and a factor of $10^{-2}$ for the L2 weight regularisation performed best. \Cref{tbl:results-BoW} illustrates \ac{MAE} results for the validation and evaluation.
A codebook size of $2500$ and assigning $500$ vectors achieves $7.85$~\ac{MAE} on evaluation.

\section{Discussion}\label{sec:discussion}
\begin{figure}[h]
\centerline{\includegraphics[width=\columnwidth]{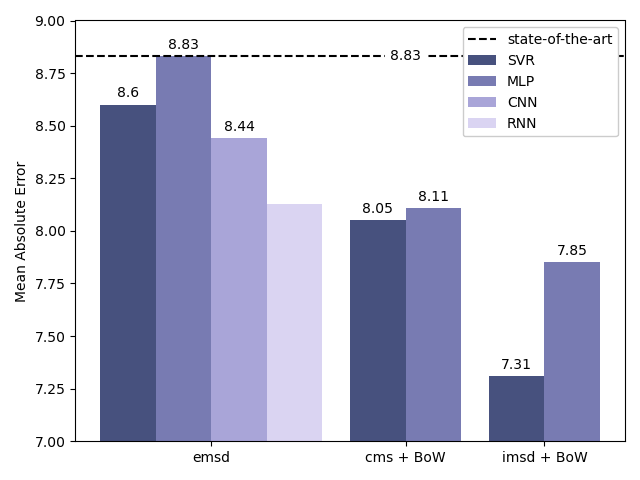}}
\caption{Best results for motility prediction and state-of-the-art~\cite{thambawita2019p61}. The lower the \acf{MAE} the better, showing that \ac{SVR} models achieve better results than the other models for every type of features. \acp{BoW} outperform models trained on \ac{emsd} features.}
\label{fig:mot-results}
\end{figure}

The large number of experiments conducted and evaluated in this article additionally require a high level overview and summary which discusses individual strengths and weaknesses. For predicting motility, almost every experiment in this article improved upon the state-of-the-art~\cite{hicks2019machine} which is already better than the ZeroR baseline. The best results of every investigated combination of feature representation and machine learning algorithm are displayed in~\Cref{fig:mot-results}. 
Using \emph{\ac{emsd}} feature vectors extracted from overlapping windows of the input videos already leads to better results with almost every machine learning model. Comparing the different algorithms for this feature type shows that more involved neural network architectures, \ie, \acp{CNN} and \acp{RNN}, are able to extract additional information from the features, by either considering structural dependencies within a single vector (CNN) or exploiting long(er) term temporal dependencies between consecutive \emph{\ac{emsd}} measurements (\ac{RNN}) -- with the latter leading to the best results with these kinds of features. However, an \ac{MLP} trained on \emph{\ac{emsd}} vectors is inferior to the more robust \ac{SVR}. The remaining three types of features all aggregate sperm-level information into subject-level sparse feature representations via \ac{BoW}. Therefore, they are no longer suitable candidates for the \ac{CNN} and \ac{RNN} models. Overall, the \ac{BoW} methodology still leads to stronger results considering the simpler machine learning strategies applied in those experiments. Furthermore, the strongest feature representations can be extracted by constructing \acp{BoW} from \emph{\ac{imsd}} vectors with the overall best \ac{MAE} of $7.31$. This result is further significantly better than state-of-the-art results at $p<.01$ measured by a one-tailed T-Test and than Hicks \etalns~\cite{hicks2019p56} at $p<.05$. For these experiments, the \ac{SVR} outperformed the \ac{MLP} and therefore, the latter is not applied in the very last experiment. This observation can be explained by the small size of the training dataset in the \ac{BoW} experiments where features are aggregated per patient. As Deep Learning models generally require larger amounts of data to perform well, this circumstance might have prevented the \ac{MLP} from achieving better results.

\section{Conclusion and Future Work}\label{sec:conclusion}
In this article, the task of automatic sperm quality assessment from microscopic video recordings is addressed by applying a framework of unsupervised tracking, feature quantisation and machine learning. The publicly available \emph{visem} dataset served as the basis for predicting the motility of spermatozoa. Two different tracking algorithms are utilised in order to enable extraction of features on a per sperm cell basis. The features are then quantised and aggregated with a \ac{BoW} approach and used as input for machine learning models. All methods herein achieved improvements for motility prediction over the submissions to the Medico Multimedia for Medicine challenge. The overall best results are achieved by unsupervised tracking of sperm cells with the Crocker-Grier~\cite{crocker1996methods} algorithm, extracting \emph{\ac{imsd}} features for each detected track and aggregating those features into a histogram representation using \ac{BoW}. With this feature representation, a linear \ac{SVR} improved the mean (3-fold) \ac{MAE} from $8.83$ to $7.31$, a decrease of over $17\,\%$. The results further show that the unsupervised feature quantisation helps to achieve more consistent and robust results, regardless of which feature representation is chosen as input. 
For future work, the presented framework can be extended and improved upon by pursuing a number of additional research directions. First of all, other methods of feature extraction from sperm tracks can be explored. During the experiments for this article, a more involved and computationally heavy set of features in the form of cell motility parameters, such as curve linear velocities and coefficients obtained from regression analysis, are evaluated. Combined with the \ac{BoW} feature quantisation however, these are less successful than the simpler \emph{\ac{imsd}} vectors. More interesting could be to integrate unsupervised representation learning into the process. A direct approach, for example, could train an autoencoder directly on the video content. Considering the noisy nature of the sperm sample recordings which contain lots of debris and background contrast variation, and furthermore exhibit very sparse motion characteristics, this seems hardly feasible with state-of-the-art deep learning methods. Instead, convolutional and recurrent autoencoders could be applied to suitable transformations of the detected tracks, as has already been done for single tracks of myogenic cells~\cite{kimmel2019deep}. Here, all tracks could be considered together or individually in an unsupervised training procedure. Using \textsc{motilitAI}, our low-resource AI-based method for automatic sperm motility recognition, we hope for its integration in digital microscopes and making our solution reachable for everyone at low cost.

\section*{Acknowledgement}
This research was partially supported by Zentrales Innovationsprogramm Mittelstand (ZIM) under grant agreement No.\ 16KN069455 (KIRun), Deutsche Forschungsgemeinschaft (DFG) under grant agreement No.\ 421613952 (ParaStiChaD), and SCHU2508/12-1 (Leader Humor).

 \bibliographystyle{elsarticle-num} 
 \bibliography{main}

\begin{acronym}
\acro{AI}[AI]{Artificial Intelligence}
\acro{BoW}[BoW]{Bag-of-Words}
\acrodefplural{BoW}[BoWs]{Bags-of-Words}
\acro{CASA}[CASA]{Computer-aided Sperm Analysis}
\acro{cms}[cms]{custom movement statistics}
\acro{CNN}[CNN]{Con\-vo\-lu\-tion\-al Neural Network}
\acrodefplural{CNN}[CNNs]{Con\-vo\-lu\-tion\-al Neural Networks}
\acro{DL}[DL]{Deep Learning}
\acro{ELU}[ELU]{Exponential Linear Unit}
\acrodefplural{ELU}[ELUs]{Exponential Linear Units}
\acro{emsd}[emsd]{ensemble mean squared displacement of all particles}
\acro{fps}[fps]{Frames per Second}
\acro{GRU}[GRU]{Gated Recurrent Unit}
\acrodefplural{GRU}[GRUs]{Gated Recurrent Units}
\acro{HSMA-DS}[HSMA-DS]{Human Sperm Morphology Analysis Dataset}
\acro{imsd}[imsd]{mean squared displacement of each particle}
\acro{LSTM}[LSTM]{Long Short-Term Memory network}
\acrodefplural{LSTM}[LSTMs]{Long Short-Term Memory networks}
\acro{MAE}[MAE]{Mean Absolute Error}
\acrodefplural{MAE}[MAEs]{Mean Absolute Errors}
\acro{MHSMA}[MHSMA]{Modified Human Sperm Morphology Analysis Dataset}
\acro{MLP}[MLP]{Multilayer Perceptron}
\acro{MSD}[MSD]{Mean Squared Displacement}
\acro{MSE}[MSE]{Mean Squared Error}
\acro{ReLU}[ReLU]{Rectified Linear Unit}
\acro{RMSE}[RMSE]{Root Mean Square Error}
\acrodefplural{RMSE}[RMSEs]{Root Mean Square Errors}
\acro{RNN}[RNN]{Recurrent Neural Network}
\acrodefplural{RNN}[RNNs]{Recurrent Neural Networks}
\acro{SOTA}[SOTA]{state-of-the-art}
\acro{SVR}[SVR]{Support Vector Regressor}
\acro{tanh}[tanh]{hyperbolic tangent}
\acro{tf-idf}[tf-idf]{term frequency-inverse document frequency}
\acro{WHO}[WHO]{World Health Organization}
\end{acronym}

\end{document}